# Socially Assistive Robot: A Technological Approach to Emotional Support

Leanne Oon Hui Yee[1], Fun Siew Sui[1], Thit Sar Zin[1], Zar Nie Aung[1], Teoh Jiehan[1] and Yap Kian Meng[1]

[1] *Research Centre for Human-Machine Collaboration, Department of Computing and Information Systems, School of Engineering and Technology, Sunway University, Bandar Sunway, Petaling Jaya 47500, Malaysia*
(Email: leanneoon@gmail.com, fsiewsui@gmail.com, thitsarzin8@gmail.com, zarzukyi3@gmail.com, 23035264@imail.sunway.edu.my, kmyap@sunway.edu.my)

**Abstract ---** In today's high-pressure and isolated society, the demand for emotional support has surged, necessitating innovative solutions. Socially Assistive Robots (SARs) offer a technological approach to providing emotional assistance by leveraging advanced robotics, artificial intelligence, and sensor technologies. This study explores the development of an emotional support robot designed to detect and respond to human emotions, particularly sadness, through facial recognition and gesture analysis. Utilising the Lego Mindstorms Robotic Kit, Raspberry Pi 4, and various Python libraries, the robot is capable of delivering empathetic interactions, including comforting hugs and AI-generated conversations. Experimental findings highlight the robot's effective facial recognition accuracy, user interaction, and hug feedback mechanisms. These results demonstrate the feasibility of using SARs for emotional support, showcasing their potential features and functions. This research underscores the promise of SARs in providing innovative emotional assistance and enhancing human-robot interaction.

## 1. INTRODUCTION

In a society increasingly characterised by stress, pressure, and isolation, the demand for emotional support has never been higher. Emotional support, defined as the act of showing care and compassion, plays a fundamental role in fostering mental well-being [1]. Research by Infurna & Luthar [2] demonstrates that strong social support networks predicted greater resilience in the face of stressful life events.

Socially Assistive Robots (SARs) are an emerging and rapidly advancing field within robotics that offers social and emotional assistance [3]. SARs emphasise human-robot interaction by implementing modern technologies including robotics, artificial intelligence, and sensors. These robots have abilities for speech recognition, facial expression analysis, and gesture recognition [3]. These capabilities allow SARs to observe and interpret human behaviour and emotions, potentially making them ideal companions that provide emotional support while also improving our overall well-being.

In today's high-pressure environments, individuals frequently encounter high levels of stress and sadness. If not properly addressed, these emotions can escalate into serious mental health conditions, such as depression [4]. Despite the prevalence of such feelings, the stigma and hesitation surrounding mental health discussion remain significant barriers to seeking help [5]. This often leads to individuals avoiding emotional support and becoming reluctant to express their feelings, thereby hindering them to manage their mental well-being effectively.

Innovative approaches such as emotional support robots, offer new forms of assistance beyond traditional methods. This project seeks to explore the potential of robots in delivering effective emotional support through technology to complement human companionship. The goal is to develop a user-friendly robot that creates a safe and supportive environment for individuals to share their emotions openly and comfortably.

## 2. RELATED WORK

Lovot, created by GROOVE X, is a social robot designed as a pet-like companion for families. It features expressive eyes, smooth fur, touch-responsive sensors, and advanced AI that recognises and reacts to user behaviour. Powered by a deep learning FPGA decision engine, it delivers lifelike movements and adapts to its environment with minimal lag. It evolves a unique personality based on continuous user interaction through behaviour learning [6].

Jibo, developed by Jibo Inc., is an AI-powered tabletop robot designed as a friendly companion, offering basic assistance such as weather forecasting [7] [8]. It is equipped with facial, voice, and speech recognition, and interacts with users through lifelike behaviours, including a youthful personality, blinking, dancing, playing games, proactive engagement [7] [8]. These features make Jibo popular for emotional coaching and learning activities in various experiments, as users often view it as a non-judgemental, pet-like companion [8].

**Liku,** from Torooc, is a humanoid robot designed to provide emotional support and companionship, particularly for lonely senior citizens adjusting to digital services. Liku uses advanced facial recognition and machine learning to identify individuals, analyse their emotions, and respond appropriately [9]. Liku has a full-colour LED screen displaying human-like facial expressions, articulated limbs for synchronised movements, and advanced voice recognition, providing empathetic and personalised interactions.

This robot integrates the strengths of Jibo, Lovot, and Liku to offer a more comprehensive user experience. Jibo excels in conversational companionship but lacks emotional adaptability due to pre-programmed responses. Lovot and Liku provide non-verbal emotional support, with Lovot's tactile design and touch-responsive sensors for emotional bonding, and Liku's lifelike gestures, though both lack conversational emotional analysis capabilities. In contrast, this robot merges physical comfort, engaging conversations and real-time emotion detection with generative AI, offering a more personalised and adaptive emotional support experience.

Table.1 Technical Comparison of Robots

| Robot | Interaction Style | Emotional Detection | Speech Recognition | Facial Recognition | Adaptive Learning | AI Capabilities |
|---|---|---|---|---|---|---|
| Lovot | Non-verbal | Basic, based on interaction | Limited, responds to touch | None | Learns from interaction | Deep learning for personality development |
| Jibo | Verbal | None | Wake-word detection, basic conversation | Recognizes multiple users | Pre-programmed responses | Limited, Pre-programmed AI |
| Liku | Non-verbal | Basic emotion analysis | Advance speech commands | Recognizes emotions and faces | Learns from user emotions | Machine learning for emotion recognition |
| Our robot | Combined | Real-time emotion detection | Advanced NLP with generative AI | Recognizes emotions and face gestures | Real-time adaptive learning | AI-generated conversation |

## 3. METHODOLOGY

### 3.1 Hardware

For this research, a basic robot is constructed using the Lego Mindstorms Robotic Kit and a Raspberry Pi4 microprocessor. **Lego Mindstorms** was selected for its versatility in creating various robots, using components such as Ev3 intelligence bricks to test the robot's functionalities in a simulated environment, servo motors to control arm movements to enable hugging, and ultrasonic sensors to detect user proximity from the robot to ensure accurate hug delivery when the user is close by. **Raspberry Pi 4** serves as the central processing unit, supporting Python for programming and providing input and output pins for accessing and processing with IoT equipment. Lastly, the robot is equipped with a **Logitech C170 webcam,** a **microphone,** and a **Bluetooth speaker** for simple real-time input and output operations to facilitate user interaction.

### 3.2 Software

**Python Idles** is used to program the system due to its compatibility with the Raspberry Pi 4, which has an inbuilt Python IDLE. The project employs six Python libraries to enhance the robot's interaction capabilities which are OpenCV, DeepFace, os module, speech-to-text, MM-LLM, and text-to-speech.

Facial recognition uses **OpenCV** (Open-Source Computer Vision Library) with Haar cascades to detect faces within images and LBPH (Local Binary Patterns Histograms) to recognise these faces based on their unique features. OpenCV also supports gesture recognition by tracking facial landmarks and using the Lucas-Kanade optical flow method to analyse motion across frames, allowing the robot to distinguish vertical movements (nods) and horizontal movements (shakes). **DeepFace** analyses facial features and recognises emotions using deep convolutional neural network (CNN) to train on extensive image datasets and categorise faces [10]. The DeepFace.analyze function filters and identifies the most dominant emotion on a scanned face.

The **OS module** manages audio playback via the MPG123 library, allowing the robot to play MP3 files through its speaker. For speech recognition, the robot relies on Python's SpeechRecognition and PyAudio packages. PyAudio captures speech through the built-in microphone, while SpeechRecognition converts the human-speech input into processible lines of string [11]. Among several available APIs, the Google Web Speech

API was chosen due to its open-source nature.

**Multimodal Large Language Model** (MM-LLM) via Groq API is integrated to generate accurate, engaging and context-aware conversations with users. Lastly, Python's pyttsx3 library was integrated into the robot's code to convert the AI-generated text into speech, ensuring seamless verbal communication with the user.

### 3.3 Testing Method

A public testing was conducted with 25 participants to assess the functionality and effectiveness of the social assistive robot through convenience sampling approach. The robot was deployed in a controlled environment with stable internet, quiet surroundings, and adequate lighting to ensure consistent testing conditions and minimise external interference. Before interacting with the robot, participants were asked to recall a recent stressful or sad experience to simulate an emotionally relevant context, assessing the robot's ability to provide emotional support in situations reflective of real-world use. Following the interaction, participants completed a survey evaluating the robot's functionality, user satisfaction, and perceived emotional support, providing key insights into its overall effectiveness and guiding potential areas for improvement.

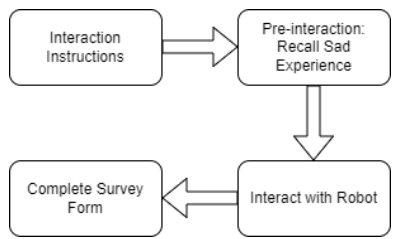

Fig.1 Testing Procedure

## 4. RESULTS/DISCUSSION

### 4.1 Experimental Evaluation

Controlled testing may have produced more favourable outcomes than real-world scenarios. Despite this, the findings offered valuable insights into the robot's functionality in human-robot interaction, user experience and physical comfort. The questionnaire results were quantitatively analysed to evaluate its performance.

In the human-robot interaction evaluation, most participants found the robot easy to interact with, and its emotional recognition system was generally accurate, though it occasionally missed gestures like nodding or head shaking. Responses were mostly relevant, but many participants experienced slight delays in response time, typically ranging between five to ten seconds. Repetitive responses were moderately common, detracting from the experience for some users. The results highlight the need for improvements in the speech recognition module to enhance responsiveness and accuracy.

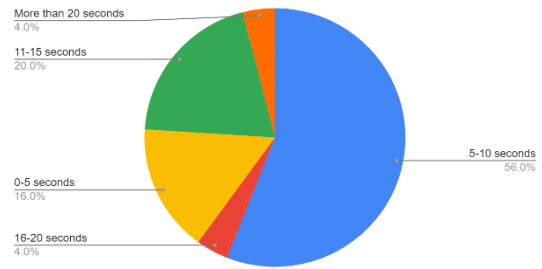

Fig.2 Robot Respond Time Distribution

In terms of user experience, participants generally found interactions with the robot engaging and smooth. The flow of conversation and response relevance were rated favourably, with most users reporting satisfaction with the robot's ability to tailor responses to their emotional cues and gestures. While some speech recognition issues occurred, requiring repetition or causing misinterpretations, the robot was largely perceived as capable of providing emotional support. Overall, users felt comfortable during interactions, indicating the system fostered ease.

Majority of the participants rated physical comfort during interactions positively, feeling relaxed with the robot's movements and proximity. The robot's design and spatial awareness contributed to a comfortable experience, engaging interactions without discomfort. The combination of intuitive interaction, emotional engagement, and physical comfort led to a generally favourable user experience, though improvements in gesture recognition and response consistency could further enhance the robot's performance.

Despite providing valuable insights, testing in controlled conditions may limit the results' applicability to real-world settings. Challenges such as speech recognition issues due to unstable internet and background noise caused delays and irrelevant responses. Therefore, broader testing across diverse users and environments is necessary for a more accurate assessment of the robot's effectiveness and reliability.

### 4.2 Comparative Analysis with related work

The robots discussed in this report were designed to provide emotional support by detecting users' emotions and offering comfort either through verbal communication, as seen with Jibo or through physical and gestural comfort, like Lovot and Liku.

The robot in this project incorporates both approaches through emotional detection with empathetic, context-specific verbal support, physical comfort though hugging, and snack provision. It surpasses Lovot and Liku by integrating Jibo's conversational capabilities and accepting both verbal and gestural inputs for a personalised interaction that learns from past interactions. Unlike Jibo's youthful personality, this robot offers a comforting and nurturing experience that excels in recognising non-verbal cues like nodding when users are too distressed to speak, surpassing both Jibo's speech-based inputs and Lovot's tactile responses [6][8].

However, the robot lacks the comforting physical features of Lovot's soft fur and Jibo's expressive voice and eyes, making it appear more robotic and less engaging. It also struggles with speech recognition in noisy environments and unstable internet, resulting in longer response times and inaccuracies. To improve engagement and effectiveness, enhancing its speech processing capabilities and physical design is necessary to improve its life-like and personable qualities.

## 5. Conclusion

This project demonstrated the potential of social assistive robots to improve emotional well-being through advanced technologies such as gesture and emotion detection, natural language processing, and conversational AI for empathetic, personalised interactions. While controlled testing showcased the robot's strengths, participant feedback identified areas for improvement, particularly in response speed and speech processing. Enhancing the robot's physical expressiveness and expanding testing to diverse real-world scenarios will be crucial for future iterations, guiding the development of a more engaging and supportive companion.

## Acknowledgement

This research is supported by the Sunway Research Centre for Human-Machine Collaboration. We extend our gratitude to Lim Zheng Jie, Alexander Ohenana Kojo Takyi Hagan, Tay Jun Sheng, and Herrick Yeap Han Lin for their invaluable insights, technical assistance, and guidance which greatly enhanced the quality of this project. Any errors are our own and should not tarnish the reputations of these esteemed individuals.